\documentclass[10pt,A4]{article}
\usepackage{geometry}
\usepackage[backend=biber, style=ieee, maxbibnames=99]{biblatex}
\usepackage[utf8]{inputenc}
\usepackage{amsmath}
\usepackage{graphicx}
\usepackage{subcaption}
\usepackage{adjustbox}
\usepackage{booktabs}
\usepackage[normalem]{ulem}
\usepackage{array}
\usepackage{color}
\usepackage[hidelinks]{hyperref}
\usepackage{nameref}
\usepackage{authblk}
\usepackage{pifont}

\definecolor{Gray}{gray}{.25}

\graphicspath{ {./imgs/} }
\newcolumntype{C}[1]{>{\centering\let\newline\\\arraybackslash\hspace{0pt}}p{#1}}

\title{Big Data, Tiny Targets: An Exploratory Study in Machine Learning-enhanced Detection of Microplastic from Filters}
\author[1,2]{\mbox{Paul-Tiberiu Miclea}}
\author[3]{\mbox{Martin Sboron}}
\author[1]{\mbox{Hardik Vaghasiya}}
\author[4]{\\\mbox{Hoang Thinh Nguyen}}
\author[4]{\mbox{Meet Gadara}}
\author[3,5,\ding{41}]{\mbox{Thomas Schmid}}

\affil[1]{Martin Luther University Halle-Wittenberg, ZIK Sili-Nano, Halle (Saale), Germany}
\affil[2]{Fraunhofer Center for Silicon Photovoltaics (CSP), Halle (Saale), Germany}
\affil[3]{Martin Luther University Halle-Wittenberg, Medical Faculty, Halle (Saale), Germany}
\affil[4]{Martin Luther University Halle-Wittenberg, Institute of Physics, Halle (Saale), Germany}
\affil[5]{Lancaster University Leipzig, School of Computing and Communications, Leipzig, Germany\bigskip}

\affil[\ding{41}]{Corresponding Author: thomas.schmid@medizin.uni-halle.de}

\date{} 
\begin{document}

\maketitle

\begin{abstract}
\noindent
Microplastics (MPs) are ubiquitous pollutants with demonstrated potential to impact ecosystems and human health. Their microscopic size complicates detection, classification, and removal, especially in biological and environmental samples. While techniques like optical microscopy,  Scanning Electron Microscopy (SEM), and Atomic Force Microscopy (AFM) provide a sound basis for detection, applying these approaches requires usually manual analysis and prevents efficient use in large screening studies. To this end, machine learning (ML) has emerged as a powerful tool in advancing microplastic detection. In this exploratory study, we investigate potential, limitations and future directions of advancing the detection and quantification of MP particles and fibres using a combination of SEM imaging and machine learning-based object detection. For simplicity, we focus on a filtration scenario where image backgrounds exhibit a symmetric and repetitive pattern. Our findings indicate differences in the quality of YOLO models for the given task and the relevance of optimizing preprocessing. At the same time, we identify open challenges, such as limited amounts of expert-labeled data necessary for reliable training of ML models.
\end{abstract}

\setlength\parindent{0pt}

\section{Introduction}

Microplastics (MP) are a global phenomenon and have been found in virtually all ecosystems worldwide. Recent studies indicate a rapid increase in microplastic pollution, doubling for example in oceans between 2008 and 2018 \cite{Bermudez2025}. Growing evidence shows that microplastics also enter human systems via inhalation, ingestion, and dermal contact \cite{Feng2023}. Microplastics have been found in human blood, lungs, placenta, and stool. While both consequences and the actual dimension of the issue are subject to ongoing research, several studies indicate pathological relevance \cite{Camia2025}. Studies suggest, for example, chronic exposure may cause oxidative stress, inflammatory responses, and potential endocrine disruption. 

MPs are often understood as polymeric particles less than 5 mm in size \cite{Arthur2009,ISO2020}, originating from primary sources (cosmetic beads) or secondary sources (degraded waste).  Solid particles are usually produced to be small, for example like microbeads in cosmetics, plastic pellets (nurdles), or synthetic fibers; they are also referred to as primary microplastics. Secondary microplastics, in contrast, result from the breakdown of larger plastic waste such as bottles, bags, or fishing nets through environmental processes (e.g. sunlight exposure, abrasion). 

While conventional detection methods (Fourier-transform infrared spectroscopy, Raman imaging, etc.) have proven good accuracy, they tend to be laborious, slow, and to require human expert analysis. To this end, material scientist have grown interest in employing machine learning (ML) to support rapid analysis and enable screening of large numbers of sample. ML, especially deep learning, has accelerated the ability to process large datasets, automating  classification and allowing to integrate real-time detection into filtration and monitoring systems.

\section{State of the Art}

\subsection{Sample Characterization Methods}

Samples containing microplastics originate from various sources. They may be collected from freshwater and marine environments using nets and sediment grabs, soil samples by isolating particles after digestion and density separation, atmospheric samples using air filters and deposition collectors, or food and bottled water using filtration and chemical analysis. For detection within the human body, samples may origin from the blood, stool, urine or biopsies \cite{barcelo2023}. Typical microscopy methods for detecting microplastics include optical microscopy, Scanning Electron Microscopy (SEM) and Atomic Force Microscopy (AFM). 

Optical microscopy represents a convenient, simple and economical tool to identify size and shape of particles and fibres. While various technologies are available for macroscopic, mesoscopic and microscopic analysis \cite{ntziachristos2010}, this approach is typically only applicable to detect microplastics with a size of 100 micrometers or more. Moreover, analyses with optical microscopy tend to be time-consuming due to the involvement of human analysts and provide the risk of judgment errors, for example, where transparent plastics are overseen \cite{primpke2022}. Likewise, natural or biological particles can lead to false identifications. While such challenges can be improved by advanced visualization using fluorescence staining which tags plastic particles, false positives remain a challenge of this approach \cite{maes2017}.

SEM and AFM provide high-resolution images of morphology and surface features, allowing characterization even at much smaller scales \cite{mariano2021}. SEM provides resolution capabilities down to 0.1 micrometers. AFM allows for analysis even down to the nanoscale without extensive sample preparation. While SEM allows to characterize the shape, texture, and surface features of microplastics similar to optical microscopy (including weathering and fragmentation patterns), the color or polymer type can not be analyzed directly, necessitating complementary spectroscopic methods \cite{huang2023}. 

Reliable screenings of both quantity and quality of particles at scale, however, remain challenging with existing methods and human analysts \cite{ivleva2021}.

\subsection{Computer Vision and Object Detection}

Modern computer vision allows to automate analysis of images in many ways. Popular deep learning approaches, such as convolutional neural networks (CNNs) and vision transformers,  allow today not only for reliable image classification systems, but even for robust object detection within and segmentation (semantic or instance) of images in various application scenarios.

While image classification represents a rather basic task in which a given image is matched with a descriptive label or category, object detection as well as segmentation are considered more challenging tasks requiring specialized models. Object detection involves identifying and localizing one or more objects within a given image by determining bounding boxes around them; often, this is combined with a classification of the objects. Image segmentation goes beyond this by basically classifying each pixel of an image; thereby, segmentation algorithms divide a given image into regions corresponding to different objects.

In the context of this study, object detection would be sufficient for the basic task of simply quantifying the number of detected particles. As we may to determined the type of particles in the long run, however, the exact shape, orientation as well as the correct location of centroids is important and requires the application of segmentation.
To this end, we employed the segmentation variants of the \textit{You Only Look Once} (YOLO) model. The choice of YOLO for this work is in particular grounded in its successful employment in a broad variety of application domains \cite{Hussain2024}, \cite{Diwan2023} even for MP detection \cite{Liu2024}, and still state-of-the-art performance compared to other models even transformer models \cite{Sapkota2025}. 

YOLO, initially introduced in 2015 by Redmon et al. \cite{Redmon2016}, is a pioneering object detection algorithm that has since become state-of-the-art for efficient object detection. Its primary innovation was the ability to predict bounding boxes and class probabilities directly from full images in a single evaluation. Later versions expanded its capabilities to areas such as pose predictions and segmentation. From Yolov5 \cite{Jocher2020}, developed by Ultralytics, onward the implementation framework was changed from Darknet to Pytorch, which made the model more accessible. Key innovations between the version include: a Feature Pyramid Network (FPN)-inspired design, which allowed for better detection across various object scales; Path Aggregation Network (PAN) to improve fusing features from different layers enhancing speed and accuracy; and Cross Stage Partial with Spatial Attention (C2PSA) module to improve spatial attention in feature maps, increasing accuracy, especially for small and overlapping objects \cite{Jegham2025}.

\subsection{Machine learning-enhanced Detection of Microplastic}

Employing ML techniques, various approaches have been developed for automating and enhancing the accuracy of the detection process \cite{huang2025}. AI-based detection tools can help quantify exposure, trace sources, and assess individual or community-level risks. Integrated sensors powered by ML can be used in water purification systems or diagnostic tools to identify MP contamination before ingestion or inhalation. A key challenge for ML-enhanced detection of microplastic remains the availability of suitable training data. While robust ML approaches usually require thousand, ten thousands or even more images for training, microplastics are not only unevenly distributed across ecosystems but also show chemically diverse compositions \cite{khanam2025}.

Applying Convolutional Neural Networks (CNNs) to detect microplastics from optical microscopy, some studies report accuracy values of more than 90 percent \cite{de2025}. Using YOLOv5 and DeepSORT, Sarker et al. (2024) achieved in situ detection of MPs in river systems, significantly enhancing ecological monitoring \cite{Liu2024}. Advanced microscopy techniques, such as Laser-induced fluorescence, have been suggested to allow for distinction between types of microplastics at 88 percent accuracy \cite{merlemis2024}. 

Combining segmentation using U-Net and
MultiResUNet with a fine-tuned VGG16, Shi et al report identification and classification of microplastics with more than 98 percent accuracy \cite{shi2022}. Together with providing a dedicated SEM data set, Rivera and colleagues applied deep-learning-based object detection algorithms (Mask R-CNN, Faster R-CNN, and YOLOv10) with likewise high accuracy in identifying and classifying microplastics \cite{rivera2025}. Groschner et al. (2021) proposed a U-Net and random forest pipeline for defect detection in TEM/SEM datasets, successfully generalizing to MP segmentation tasks with a Dice score of 0.8 \cite{Groschner2021}.

\section{Methodology}

\subsection{Data Acquisition}\label{micro-raman-spectroscopy}

\textbf{Substrate \& Filters.} The substrate used for all filtration experiments consisted of silicon (Si) filter membranes with a nominal pore size of 1 $\mu$m, produced via chemical etching by SmartMembranes GmbH (Germany). These membranes are laser-cut into a circular format with a 9 mm diameter and a thickness of approximately 300 $\mu$m. Each membrane features a square pore structure designed for effective filtration of micro- and nanoparticles.

\begin{figure}[h!]
\centering
\begin{minipage}[c]{0.47\textwidth}
\centering
\includegraphics[width=\textwidth]{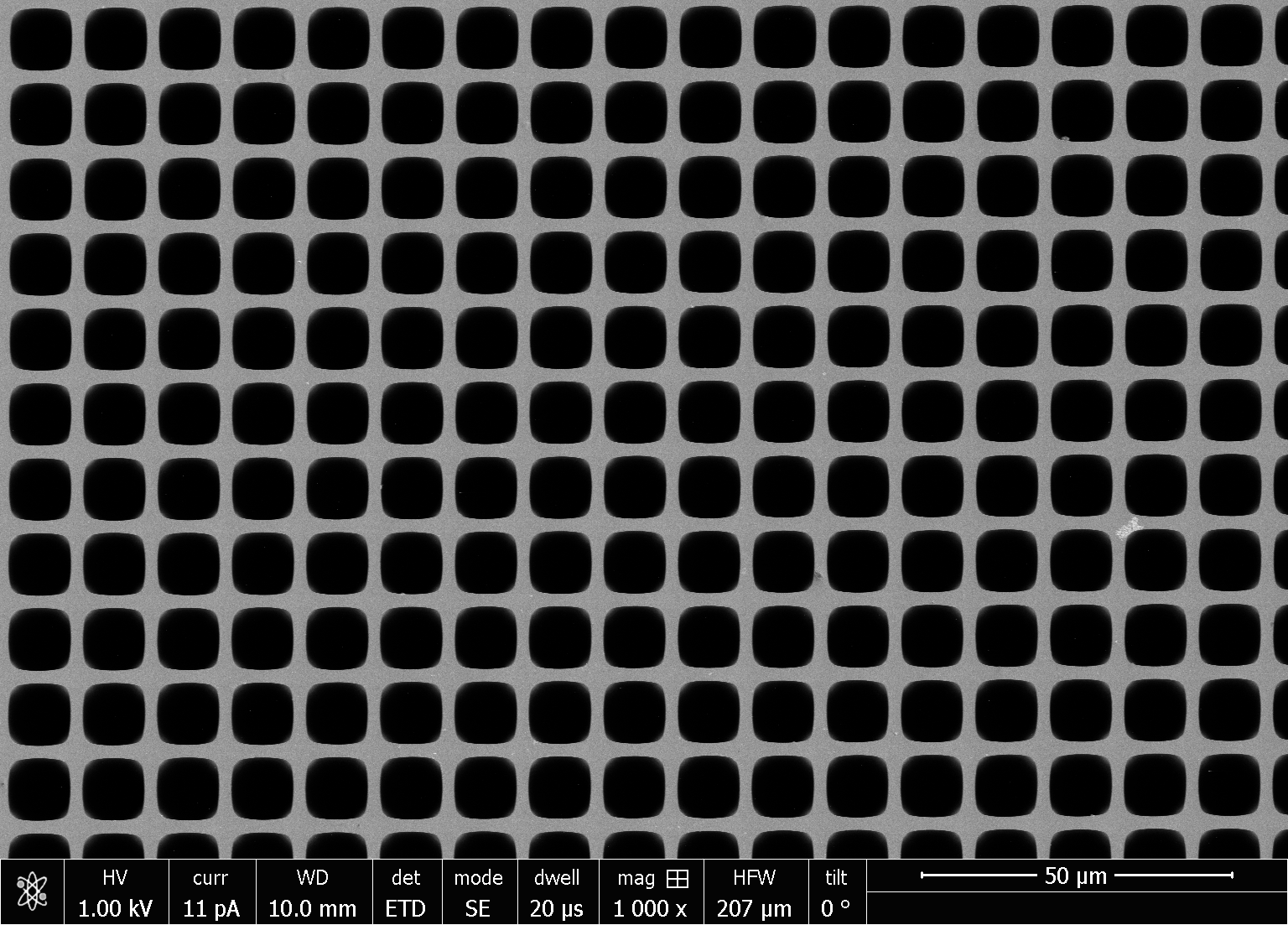}
\end{minipage}
\hspace{0.5cm}
\begin{minipage}[c]{0.47\textwidth}
\centering
\includegraphics[width=\textwidth]{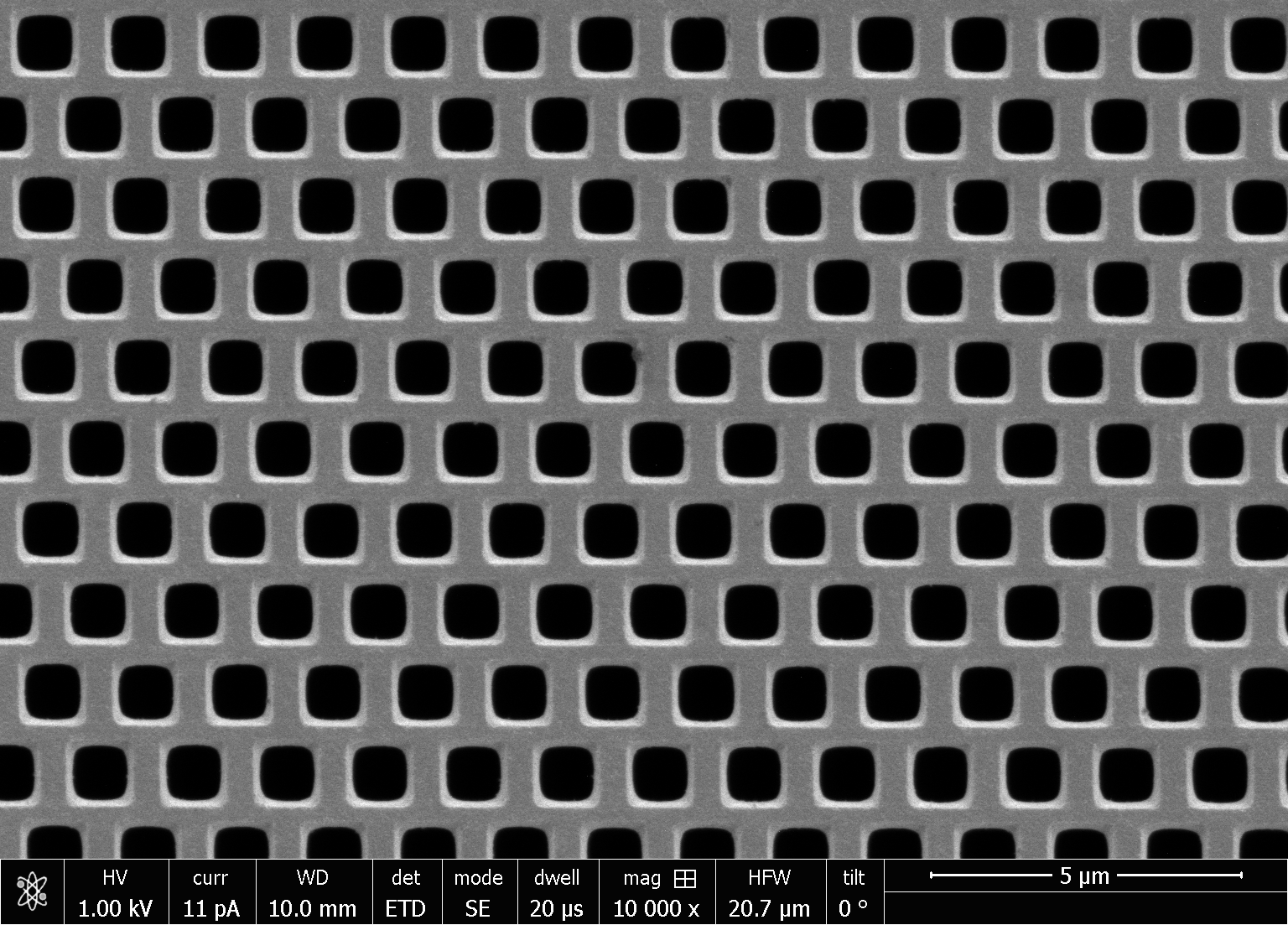}
\end{minipage}
\vspace{-0.1cm}
\caption{SEM images of the 10 $\mu$m (left) and 1 $\mu$m (right) pore size Si filter membrane }
\label{fig:filter_orig}
\end{figure}

The filters were mounted in a Sartorius filtration funnel as described in the standard operating procedure (SOP) published in Zenodo~\cite{Miclea_2025}. Unlike the reference SOP, the filtrates used in this study were analyzed in their original state without the addition of spiked microplastic particles. In general, for filtration in the cascade system a Si membrane filter with pore size of 10 $\mu$m pore size and a thickness of about 300 $\mu$m and a membrane with 1 $\mu$m pore size and a thickness of about 250 $\mu$m ad presented in Figure \ref{fig:filter_orig} was used.

\label{optical-microscopy}
\label{SEM-microscopy}

\begin{figure}[b!]
\centering
\begin{subfigure}[c]{0.3\textwidth}
\centering
\includegraphics[width=\textwidth]{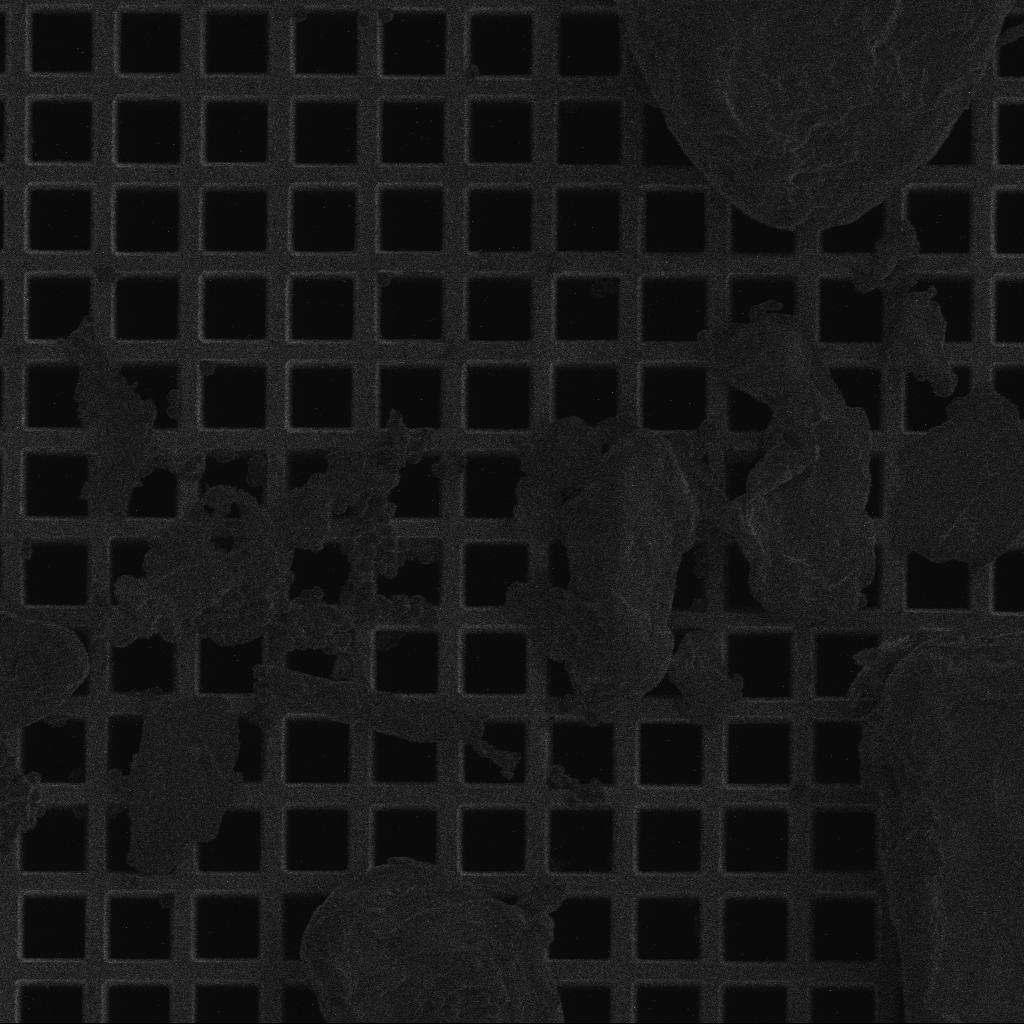}
\caption{}
\end{subfigure}
\hfill
\begin{subfigure}[c]{0.3\textwidth}
\centering
\includegraphics[width=\textwidth]{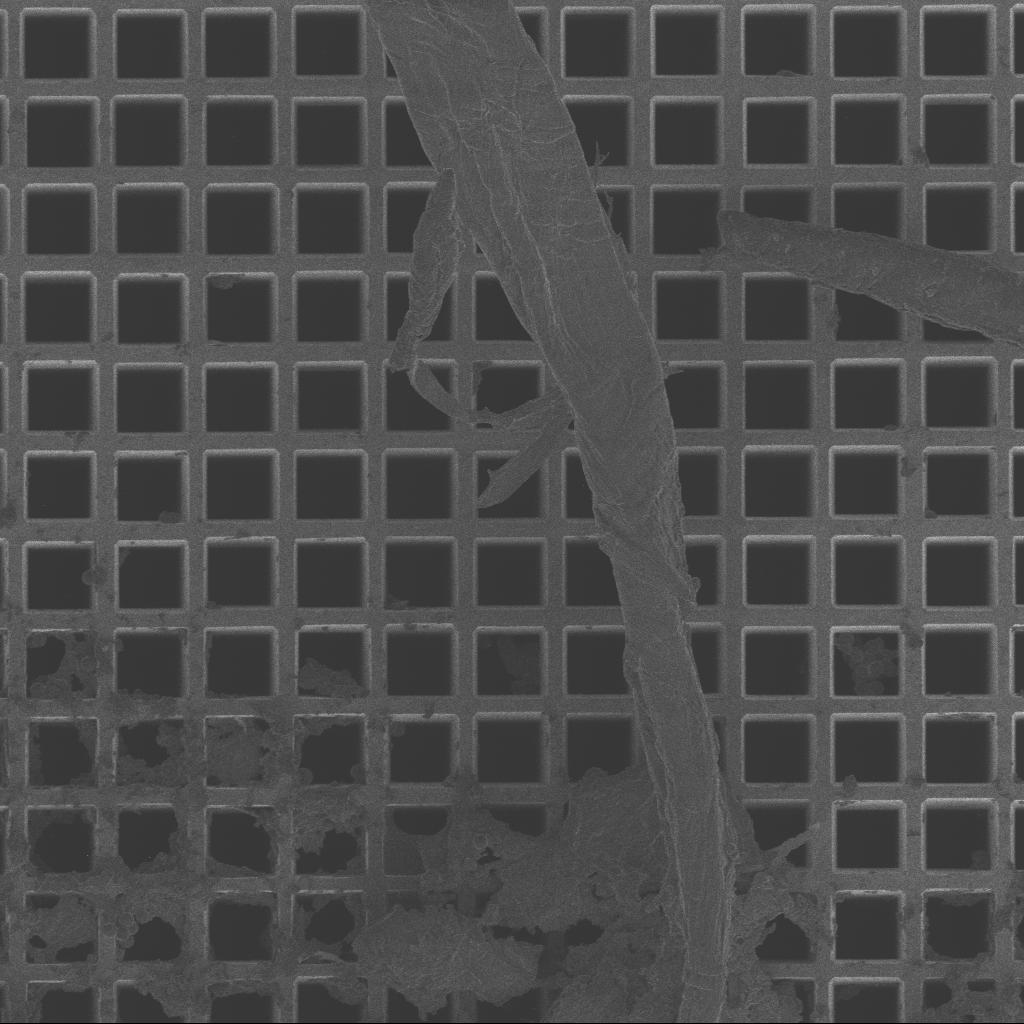}
\caption{}
\end{subfigure}
\hfill
\begin{subfigure}[c]{0.3\textwidth}
\centering
\includegraphics[width=\textwidth]{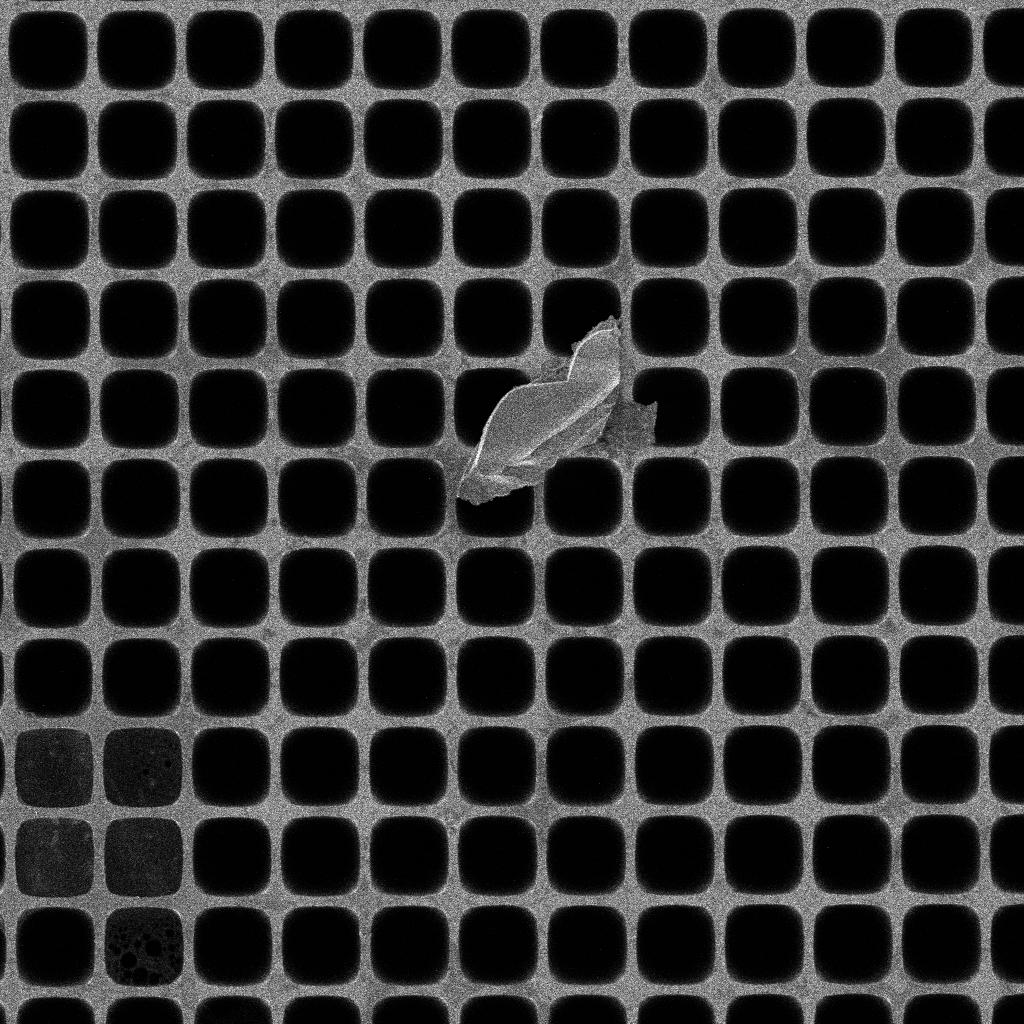}
\caption{}
\end{subfigure}
\caption{\small{Images with differing illumination and contrast with (a) low illumination, (b) low contrast, (c) good contrast and illumination}}
\label{fig:illu-diffs}
\end{figure}

\textbf{Scanning electron microscopy.} SEM was used to visualize structural features and surface textures for particle samples with sizes below the limit of detection of optical microsope. For small microplatic imaging, below the limit of detection of the optical microscope, a scanning electron microscope Versa 3D DualBeam (ThermoScientific formelly FEI) was used. In order not to contaminate the sample after filtration, no deposition of conducting layer was used. For this purpose the acceleration voltage of the microscope was chosen so it did not affect the microplastic particles. The measurement and counting of particles follow standardized procedures, in particular the directive DIN ISO 19748:2023-07, which defines requirements for particle size analysis and counting accuracy \cite{DINISO19748_2023}.

\subsection{Data Preparation \& Preprocessing}

\textbf{Labeling.} 195 of the acquired images were manually annotated to indicate microplastic particles and fibres. To support this expert-driven task, the software makesense.ai was used \cite{make-sense}. After the initial training runs with YOLOv8, additional segmentation masks predicted by the model were corrected, if necessary, and also kept as ground truth.

\textbf{Cropping.} For a smoother training of the YOLO model a symmetric data format is favorable. To this end, all images were cropped to a power of 2 size of 1024x1024 to avoid uneven padding for the model. 

\textbf{Sample Selection.}
After labeling, particles with a length or width of smaller than 60\% of the pore size diagonal of the corresponding filter were removed from the image to prevent the model from biasing towards: first, small perturbations in the images and second, small particles that should be detected much later in the filter cascade. Since the images vary slightly in their zoom-factor, presumable due to slight skewing of the filters when the images were taken, the detection of the filter pores was achieved by classical computer vision methods. First a threshold was applied to the image, then it was inverted and after that the connected components of a reasonable size for the filter pores were determined. Finally the most frequent connected component size was taken as reference for the filter pore size. Images with no segmentation masks remaining were removed from the dataset, so that an image had at least one labeled particle. After a final manual filtering to remove images with too many unlabeled particles 152 images with 1184 particles overall remained. As a result the following filter type distribution arose: Si filter 1$\mu$m: 58 images, Si filter 10$\mu$m: 36 images, Crate filter straight: 26 images, and Crate filter skewed: 32 images.

\textbf{Contrast Correction.} Acquired raw data showed a variety of contrast and illumination differences among each other (cf. Figure \ref{fig:illu-diffs}). Based on the observation that object detection under such conditions is challenging for human detectors, two existing preprocessing methods were tested separately as counteraction methods:
\begin{enumerate}
    \item \textit{Binarization.} In order to maximize the contrast by removing all gray-scale colors, we employed Otsu-thresholding \cite{Otsu1975}, in which pixels are separated into two classes: foreground and background and the threshold results from inter-class variance maximization. 
    \item \textit{Histogram equalization.} In order to achieve equal distribution of colors across the spectrum from black to white, we employed the Contrast Limited Adaptive Histogram Equalization (CLAHE) algorithm \cite{Pizer1987} with the implementation of the scikit\footnote{\url{https://scikit-image.org/docs/0.25.x/api/skimage.exposure.html\#skimage.exposure.equalize\_adapthist}} library. In Adaptive Histogram Equalization the image is divided into small regions, in which the pixels are transformed to improve local contrast based on their distributions in that area. But this method tends to overamplify noise, so in CLAHE before computing the transformation, a "clip limit" is set to cap the maximum height of the histogram bins. Bins exceeding the limit are clipped and have their clipped parts redistributed across the other bins, therefore flattening the histogram and limiting steep local contrast changes.
\end{enumerate}

\begin{figure}[t!]
\centering
\begin{subfigure}[c]{0.3\textwidth}
\centering
\includegraphics[width=\textwidth]{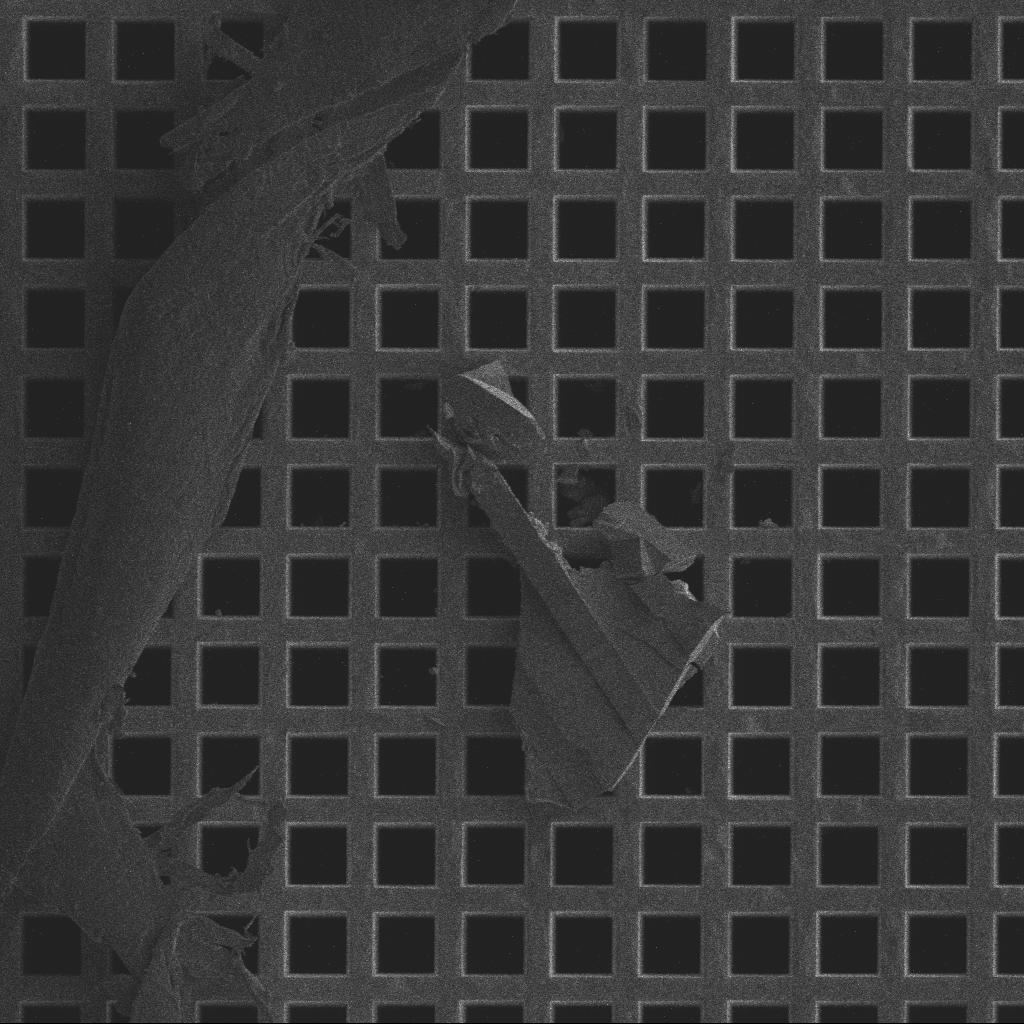}
\caption{}
\end{subfigure}
\hfill
\begin{subfigure}[c]{0.3\textwidth}
\centering
\includegraphics[width=\textwidth]{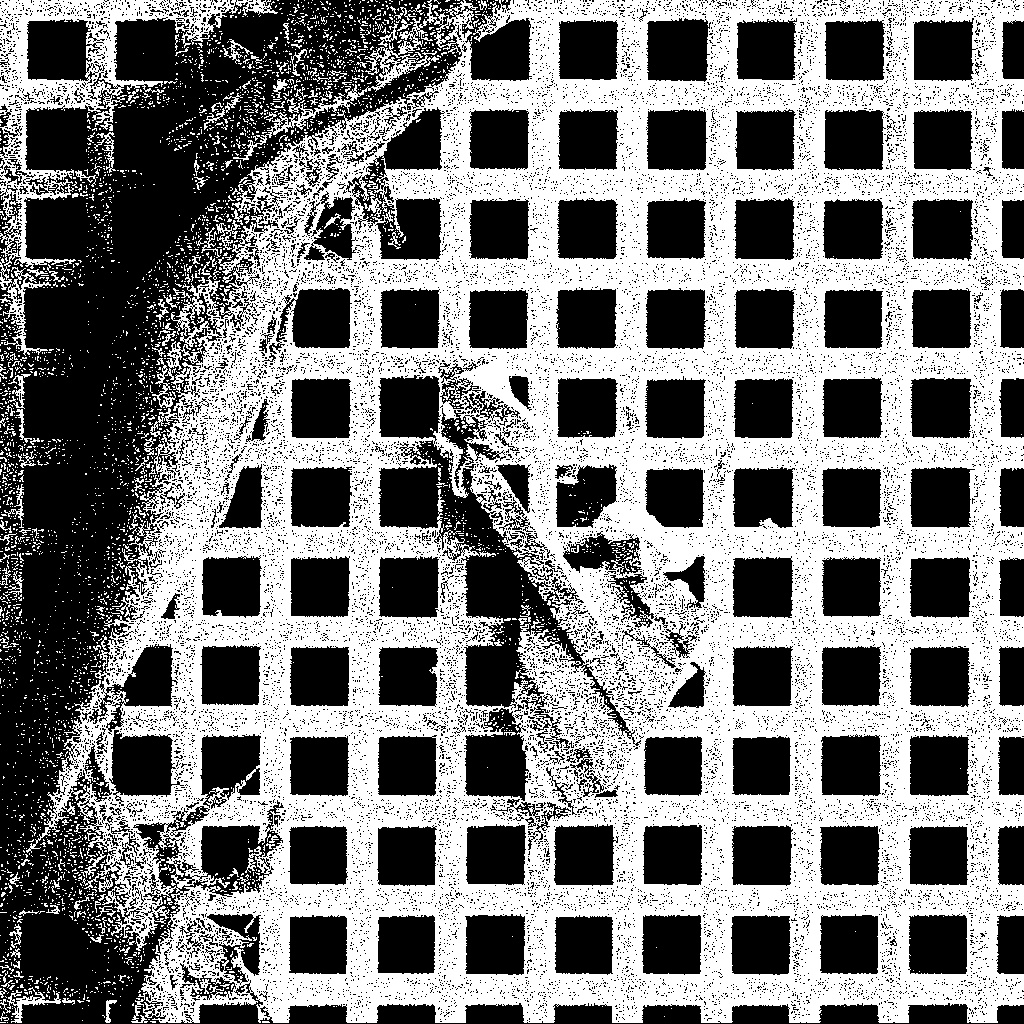}
\caption{}
\label{fig:bin}
\end{subfigure}
\hfill
\begin{subfigure}[c]{0.3\textwidth}
\centering
\includegraphics[width=\textwidth]{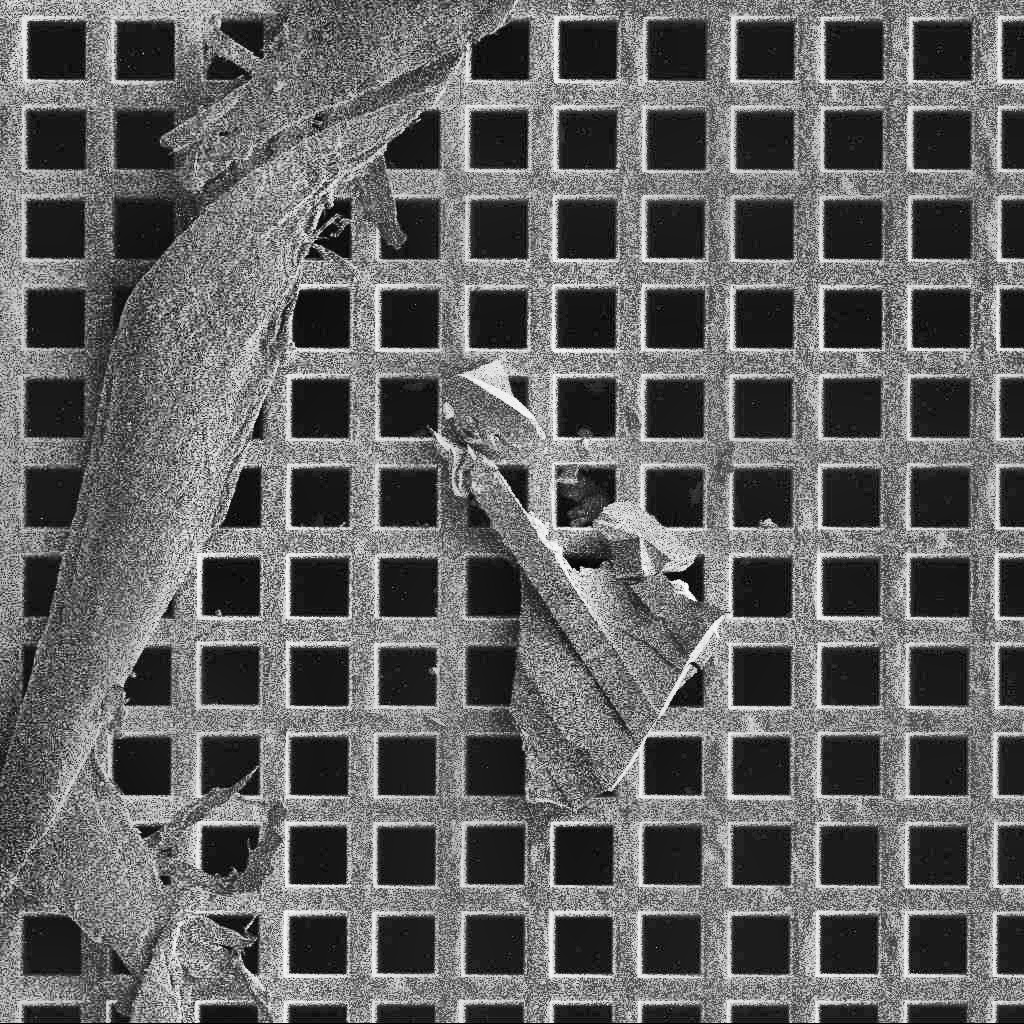}
\caption{}
\label{fig:hist-eq}
\end{subfigure}
\caption{\small{(a) original image, (b) binarization with the Otsu-method, (c) histogram equalization via CLAHE}}
\label{fig:bin-hist-eq}
\end{figure}

Both methods are illustrated in Figure \ref{fig:bin-hist-eq} on an image with originally lower illumination. It can be seen that binarization enhances contrast, but can also lead to dark area or shadow amplification, while histogram equalization provides a smoother contrast enhancement, raising even the visibility of the textures.

\subsection{Machine Learning}

\textbf{Task/model.} Originally the dataset was comprised out of two classes: particles and fiber. But since the fiber class is determined by the ratio of the length and width of a particle, which can be determined via classical methods in post-processing, the approach was reduced to a one class detection/segmentation problem. As segmentation models we compared YOLO 8, 9 and 11 from Ultralytics. YOLOv10 and YOLO12 were not considered in this work, since no segmentation versions were available at time of the experiments.

\textbf{Train-test split.} Training data was manually sorted into a 80/20 training/test split and the training split further into a 80/20 training/validation split, manually stratified by filter types. So the training set-up is: 98 training, 24 validation and 30 test images. 

\textbf{Hyperparameters.} All models were trained on the same seed and with batch-size: 4 for 250 epochs. After some initial tests batch size 4 was a good compromise between precision and recall maximization compared to sizes 8 or 16 on the given dataset. The optimizer was kept to the default, which is AdamW \cite{Loshchilov2019} with a learning rate of 2e-3, a momentum of 0.9, and a weight decay of 5e-4. Dropout was not used. All other hyperparameters were kept on the default settings from the Ultralytics trainer configuration as of their github release: v8.3.200. The best performing model on the validation set, measured by model fitness\footnote{fitness = mAP50 * 0.9 + mAP50-95 * 0.1}, was taken for the evaluation on the test set. All models were trained on an A100 GPU with training times ranging from about 0.3 to 1 GPU-hour, with larger versions of the models and especially YOLOv9 populating the higher end of the range.

\textbf{Evaluation.} Models were compared between versions of similar sizes, different sizes of the same version, and with binarization or histogram equalization with the same version and same size. The models were evaluated on the test-set, which contained 30 images with 277 particles in total with metrics precision, recall, and mean average precision across Intersection over Union for threshold 0.5 (mAP50) and thresholds from 0.5 to 0.95 (mAP50-95).

\section{Results}

\subsection{Comparison of YOLO Variants}

Table \ref{tab:size-vers-comp} shows the version and size comparison. Overall version 8n performs best except for precision, closely follow by versions 8s and 11n, which performed better in regards to precision and only lack slightly in both mAP-scores, but fall behind in terms of recall. Version 9c performs best in terms of precision but lacks quite significantly in recall and precision of the segmentation mask overlap i.e. mAP-scores compared to most of the other variants. In general, the smaller variants of the model version perform better in regards to their F1-scores. For version 8 precision improves for middle sized variants, but only for version 8s the improvement is high enough and the trade-off in the decrease of recall low enough, that the F1-score improves over the 8n version. For version 9 and 11 there is only a slight improvement in recall with increasing parameter size. All other parameters decrease with larger model variants for these two version, with one outlier of 11l for which the recall is significantly higher than all other variant of version 11, but at the price of the lowest precision over all.

\begin{table}[h!]
\centering
\small
\begin{adjustbox}{width=1\textwidth}
\begin{tabular}{c | c | C{1.8cm} | C{1.8cm} | C{1.8cm} | C{1.8cm} | C{1.8cm} } 
\toprule
Model Version & Params. & F1-Score & Precision & Recall & mAP50$\uparrow$ & mAP50-95$\uparrow$ \\ [0.5ex] 
 \midrule
8n-seg & $3.4$ & \dashuline{$0.661$} & $0.644$ & $\boldsymbol{0.678}$ & $\boldsymbol{0.678}$ & $\boldsymbol{0.468}$\\
8s-seg & $11.8$ & $\boldsymbol{0.677}$ & $0.719$ & \dashuline{$0.639$} & $\underline{0.668}$ & \dashuline{$0.457$}\\
8m-seg & $27.3$ & $0.656$ & \dashuline{$0.726$} & $0.599$ & $0.652$ & $0.441$\\
8l-seg & $46.0$ & $0.651$ & $0.685$ & $0.621$ & $0.637$ & $0.391$\\
8x-seg & $71.8$ & $0.607$ & $0.612$ & $0.603$ & $0.610$ & $0.369$\\
 \midrule
9c-seg & $27.9$ & $0.626$ & $\boldsymbol{0.742}$ & $0.542$ & $0.610$ & $0.361$\\
9e-seg & $60.5$ & $0.581$ & $0.603$ & $0.560$ & $0.571$ & $0.347$\\
 \midrule
11n-seg & $2.9$ & $\underline{0.666}$ & $\underline{0.727}$ & $0.614$ & \dashuline{$0.655$} & $\underline{0.458}$\\
11s-seg & $10.1$ & $0.644$ & $0.660$ & $0.628$ & $0.635$ & $0.447$\\
11m-seg & $22.4$ & $0.624$ & $0.653$ & $0.598$ & $0.615$ & $0.406$\\
11l-seg & $27.6$ & $0.617$ & $0.579$ & $\underline{0.661}$ & $0.627$ & $0.399$\\
11x-seg & $62.1$ & $0.585$ & $0.674$ & $0.516$ & $0.583$ & $0.361$\\
\bottomrule
\end{tabular}
\end{adjustbox}
\caption{\small{Model version and size comparison between version 8, 9 and 11. Model parameters are given in millions. Bold is best value of the metric, underlined is second best and dashed underlined is third best.}}
\label{tab:size-vers-comp}
\end{table}

\subsection{Comparison of Contrast Correction Approaches}
The best two models from the version and size comparison were then taken and compared against versions of themselves trained and evaluated with the preprocessed images, shown in Table \ref{tab:preproc-comp}. Considering F1-score as well as individual precision and recall scores in addition to the mAP-scores: models 8n and 11n showed the best performance and were chosen. The two models react differently when trained on each preprocessing method. For version 8n precision improves and recall decreases for both methods, while mAP-scores decrease for binarization, but hardly change for histogram equalization. Version 11n on the other hand shows the best precision for both model versions over all methods, with similar recall to its version without preprocessing. Binarization shows better recall than the histogram equalization method for version 11n, but lacks in precision and mAP-scores.

\begin{table}[h!]
\centering
\small
\begin{adjustbox}{width=1\textwidth}
\begin{tabular}{c | c | C{1.8cm} | C{1.8cm} | C{1.8cm} | C{1.8cm} | C{1.8cm} } 
\toprule
Model Version & Pre-proc. & F1-Score & Precision & Recall & mAP50$\uparrow$ & mAP50-95$\uparrow$ \\ [0.5ex] 
 \midrule
8n-seg & - & $0.661$ & $0.644$ & $\boldsymbol{0.678}$ & $\boldsymbol{0.678}$ & $\boldsymbol{0.468}$\\
8n-seg & B & $0.660$ & $0.720$ & $0.610$ & $0.636$ & $0.399$\\
8n-seg & H & $\underline{0.675}$ & $\underline{0.733}$ & $0.625$ & $\underline{0.670}$ & $\underline{0.460}$\\
 \midrule
11n-seg & - & $0.666$ & $0.727$ & $0.614$ & $0.655$ & $0.458$\\
11n-seg & B & $0.667$ & $0.687$ & $\underline{0.649}$ & $0.628$ & $0.385$\\
11n-seg & H & $\boldsymbol{0.684}$ & $\boldsymbol{0.767}$ & $0.617$ & $0.665$ & $0.442$\\
\bottomrule
\end{tabular}
\end{adjustbox}
\caption{\small{Model comparison between preprocessing methods: binarization (B) and histogram equalization (H). Bold is best value of the metric and underlined is second best over both versions.}}
\label{tab:preproc-comp}
\end{table}

\section{Discussion}

\subsection{Model Comparison}

\begin{figure}[b!]
\centering
\begin{subfigure}[c]{0.45\textwidth}
\centering
\includegraphics[width=\textwidth]{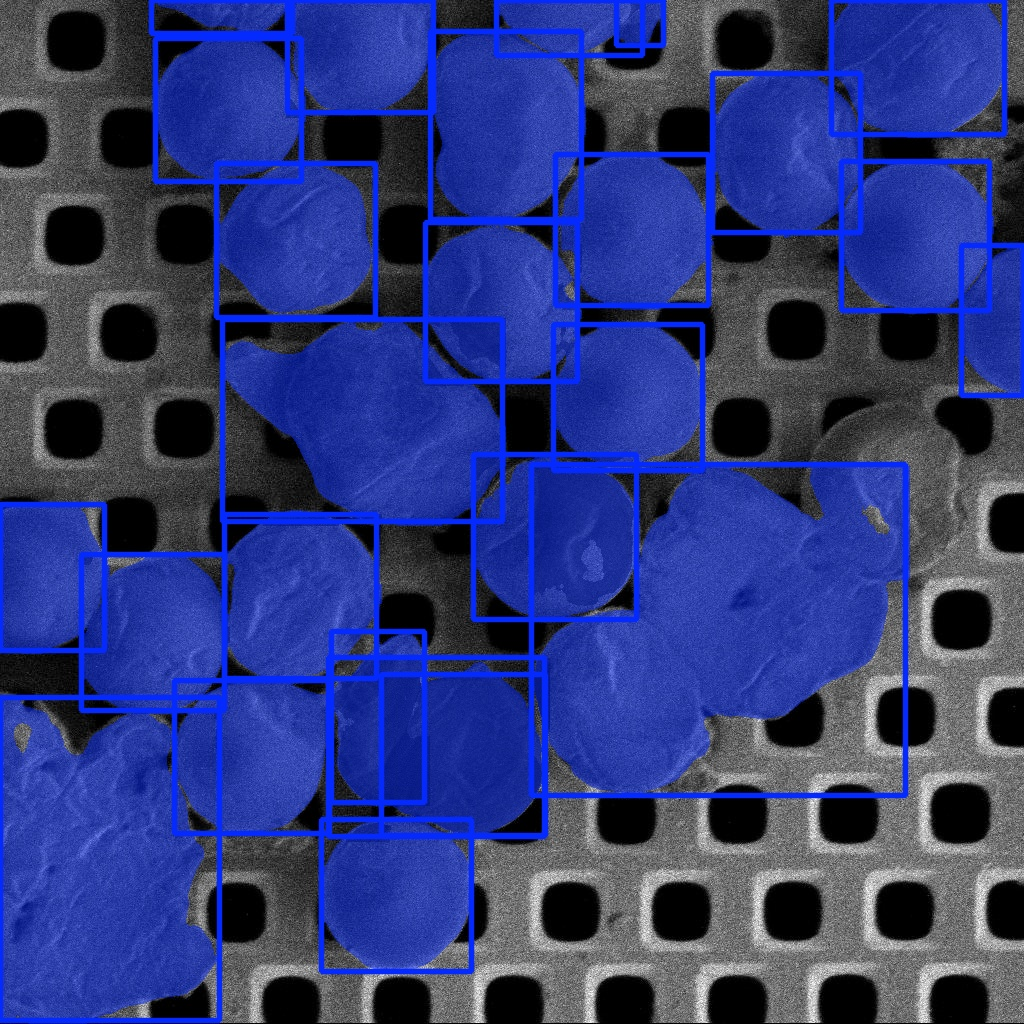}
\caption{}
\label{fig:oe-bb-11}
\end{subfigure}
\hfill
\begin{subfigure}[c]{0.45\textwidth}
\centering
\includegraphics[width=\textwidth]{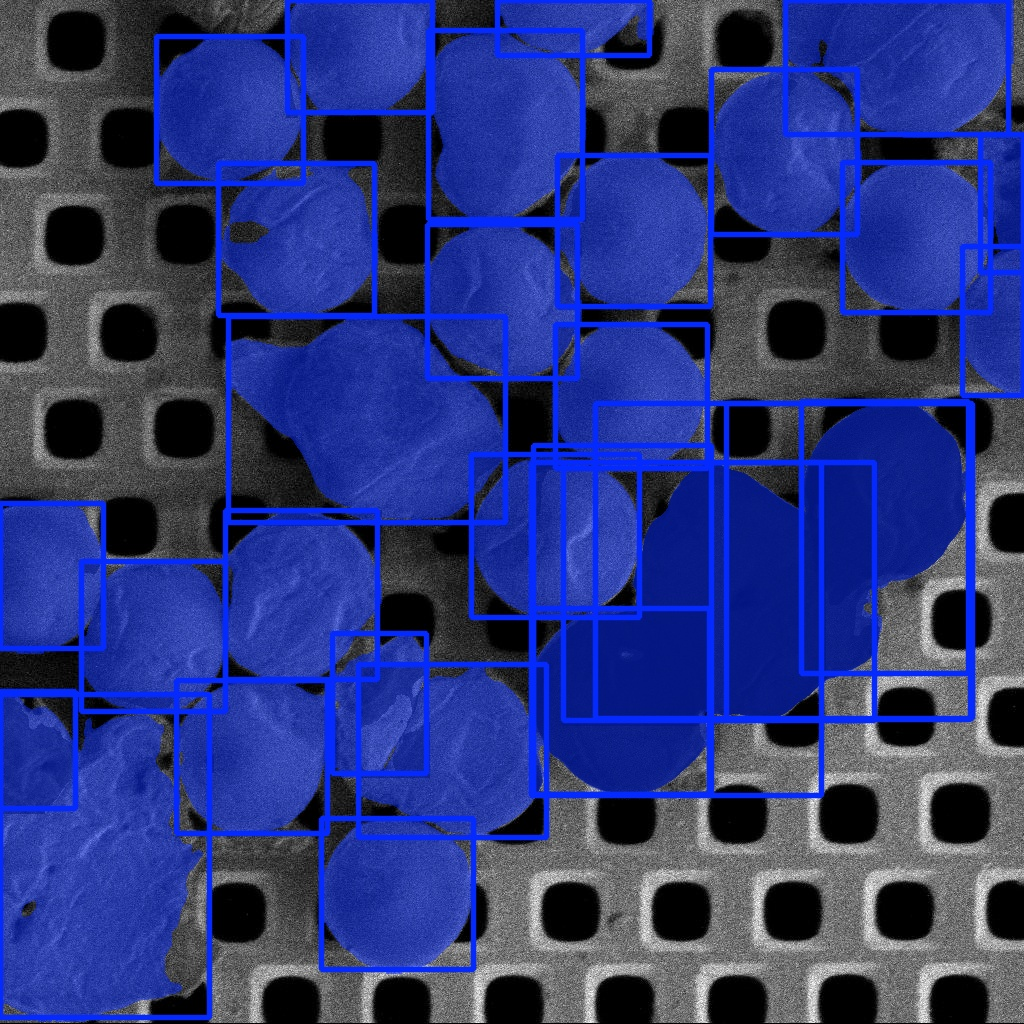}
\caption{}
\label{fig:oe-bb-8}
\end{subfigure}\\[-0.25cm]
\caption{\small{(a) segmentation mask predictions of model 8n, (b) overemphasizing of particles for the predictions of model 11n (dark blue region) of the same image.}}
\label{fig:overemp-bbs}
\end{figure}

\begin{figure}[h!]
\centering
\begin{subfigure}[c]{0.45\textwidth}
\centering
\includegraphics[width=\textwidth]{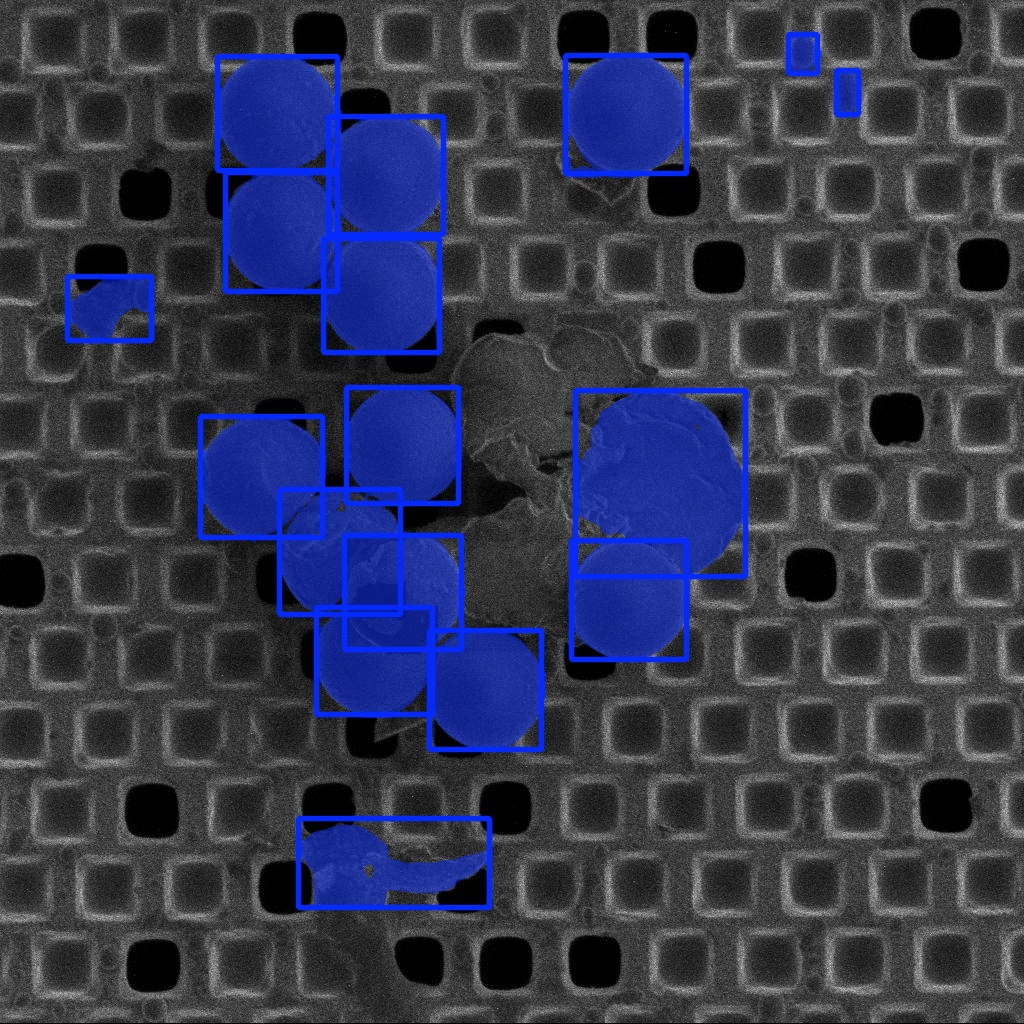}
\caption{}
\end{subfigure}
\hfill
\begin{subfigure}[c]{0.45\textwidth}
\centering
\includegraphics[width=\textwidth]{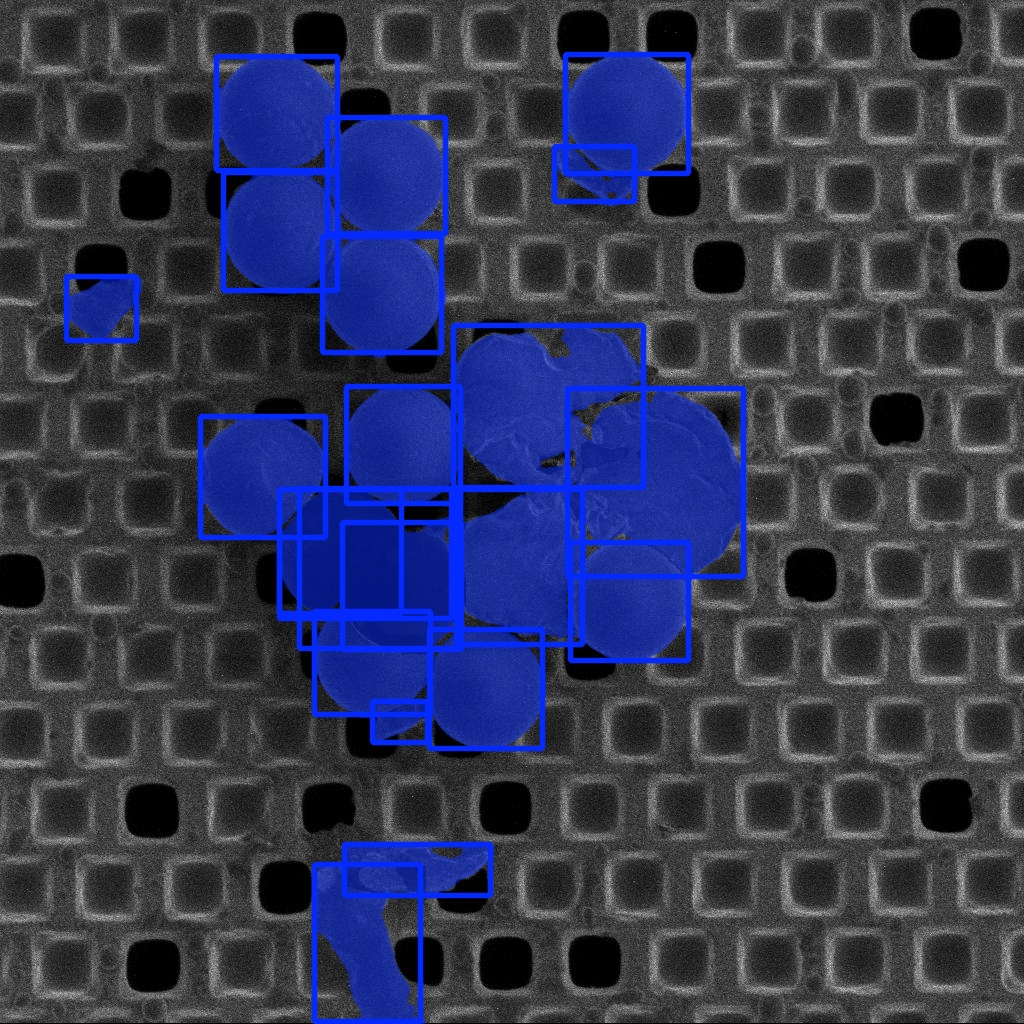}
\caption{}
\end{subfigure}\\[-0.25cm]
\caption{\small{(a) model 8n: segmentation mask missing in the middle, (b) model 11n: the same image with the full segmentation mask}}
\label{fig:seg-diffs}
\end{figure}

Overall, both the YOLO 8n and the YOLO 11n model seem to perform well on the test data. Compared to the 8n model, the 11n model tends to overemphasize some regions and separate larger objects into smaller ones compared to 8n, as can be seen in Figure \ref{fig:oe-bb-11}. On the other hand for images with overlapping particles or many particles sharing a small space 11n has better recall, Figure \ref{fig:seg-diffs} shows an example. Overall 11n also suffers from inaccurate segmentation masks and tends to miss small parts of particle, which add up over all images. But the part it misses often differ to the part 8n misses, so no clear trend is visible.

\begin{figure}[b!]
\centering
\begin{subfigure}[c]{0.45\textwidth}
\centering
\includegraphics[width=\textwidth]{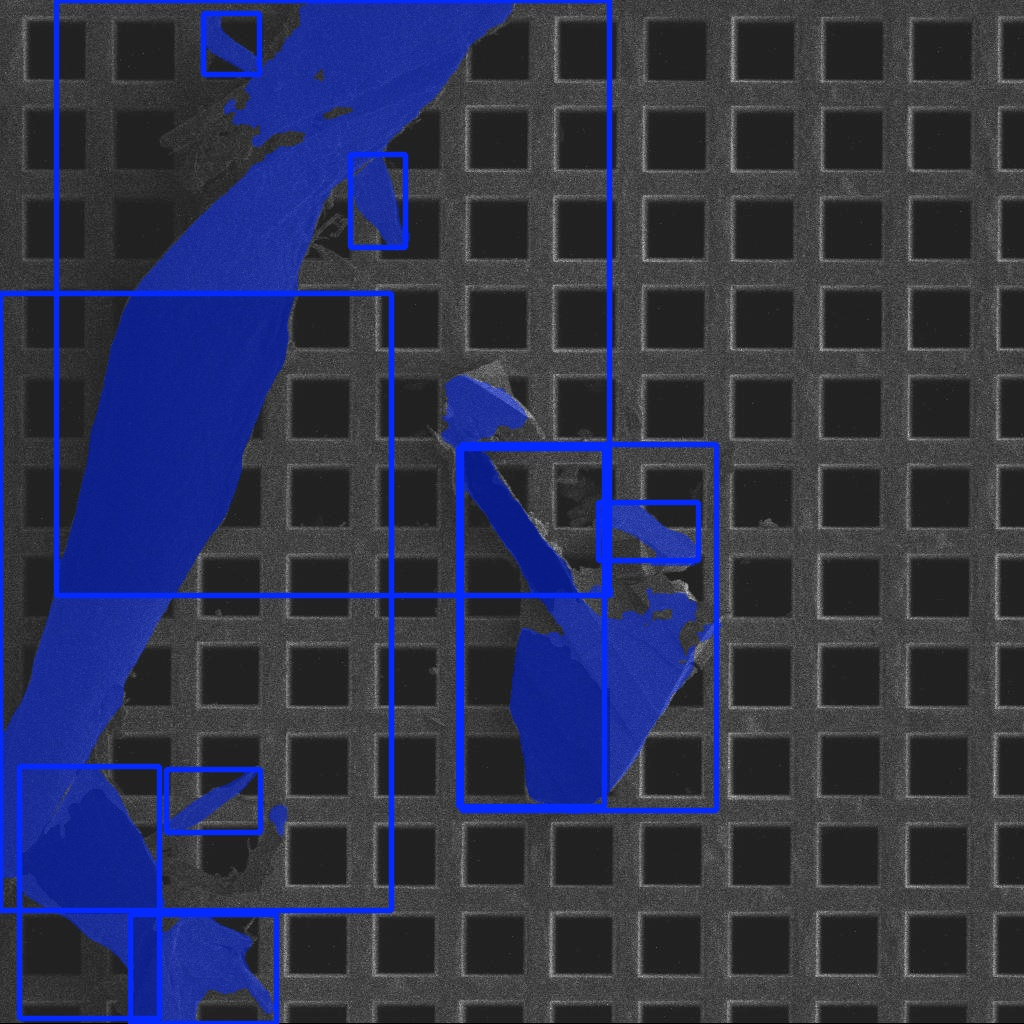}
\caption{}
\end{subfigure}
\hfill
\begin{subfigure}[c]{0.45\textwidth}
\centering
\includegraphics[width=\textwidth]{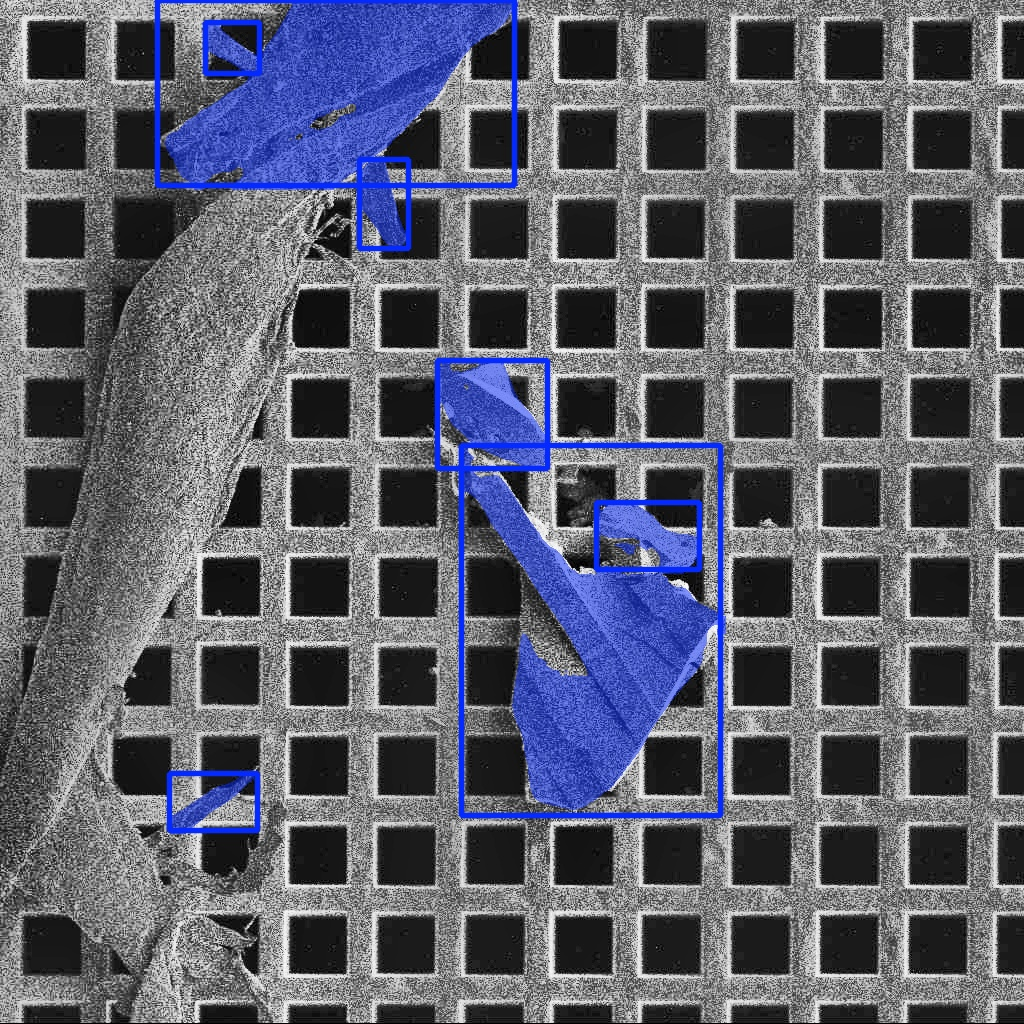}
\caption{}
\label{fig:preproc-lim-he}
\end{subfigure}\\[-0.25cm]
\caption{\small{(a) predicted segmentation masks from model 11n without preprocessing, (b) same image preprocessed with histogram equalization with missing parts of the particle in the predicted segmentation mask (also for model 11n).}}
\label{fig:preproc-lim}
\end{figure}

\subsection{Impact of Preprocessing Variants}

Considering the F1-scores of the version that were tested on the preprocessing methods binarization generally seem to trade recall for precision and does not bring much benefit overall. For histogram equalization the model version makes a difference: while for both versions precision increases, recall only significantly drops for version 8n. But even version 11n misses partly large portions of particle in some images, but   over all images the recall score amortizes close to the score of the version trained on the unprocessed images. One of these misses can be seen in Figure \ref{fig:oe-bb-11}. Since convolutional neural nets are known to concentrate on local features and can therefore have difficulties identifying large objects, one possible explanation can be that the texture of the particle is overemphasized through the preprocessing and looks locally very similar to the metal structure of the filter and therefore was misclassified as background.

\subsection{Challenges \& Current Limitations}

Our exploratory study has provided initial insight into the specific domain and revealed domain-specific challenges. Limitations in general arise when SEM images are quite cluttered or when single particles fill large parts of the image. Especially in these cases segmentation mask have smaller outlines than the particle or miss parts of the particle. False positives rates tend to be higher if partial particles are shown on the edges of the image, which are not part of the ground truth labeling, if the potential centroid of the particle is not visible. Future work may benefit from annotation of these partial particle as a separate class, in order to decrease model confusing of what are valid particles.

\section{Conclusions \& Outlook}

In this exploratory study, we have investigated potential, limitations and future directions of advancing the detection and quantification of MP particles and fibres using a combination of SEM and machine learning-based object detection. In particular, we demonstrated that both the choice of the detection model as well as preprocessing has direct impact on the detection results.
 
In future work, we will extend our work and address the identified challenges, in particular regarding quantity and quality of training data. Moreover, we will investigate the methodological extensions, such as multimodal approaches incorporating additional data sources. Further, we plan to evaluate more advanced ML models, such as vision transformer models, for the given task. 

\section*{Contributions}
PTM supervised data acquisition, labeled microplastic particles and fibres on acquired data, carried out cropping, and wrote the initial draft of the manuscript. MS optimized data preprocessing, carried out the machine learning experiments and documented both methodology and results for this. HTN, HV, and MG carried out measurements and documented related procedures. TS lead the conceptualization of the machine learning methodology and revised the manuscript.

\section*{Acknowledgments}
The authors thank the European Union for funding the H2020 projects POLYRISK (964766) as well as Euromat for funding PlasticTrace (21GRD07). Thanks to Smartmembranes for providing filters for this work, Susanne Gramsall and  Steffi Goller from Fraunhofer CSP for execution of filtration experiments.

\interlinepenalty=10000

\printbibliography

\end{document}